\documentclass{article}
\usepackage{ijcai16}

\usepackage{times}
\usepackage{latexsym}
\usepackage{amsfonts}
\usepackage{algorithmicx}
\usepackage{algpseudocode}
\usepackage{algorithm}
\usepackage{graphicx}
\usepackage{array}
\usepackage{caption,subcaption}
\usepackage{amsmath, amsthm, amssymb}
\usepackage{relsize}
\usepackage{amsmath}
\usepackage{soul}

\newcommand{\RR}[0]{\mathbb{R}}
\newcommand{\LineComment}[1]{\State \(\triangleright\) #1}
\newcommand{\vect}[1]{\mathbf{#1}}
\newcommand*{\Resize}[2]{\resizebox{#1}{!}{$#2$}}%

\usepackage{etoolbox}
\makeatletter
\patchcmd{\maketitle}
 {\def\@makefnmark}
 {\def\@makefnmark{}\def\useless@macro}
 {}{}
\makeatother

\DeclareMathOperator*{\argmax}{\arg\!\max}

\title{Distraction-Based Neural Networks for
Document Summarization\thanks{Published in IJCAI-2016: the 25th International Joint Conference on Artificial Intelligence.}}
\author{Qian Chen$^\dagger$, Xiaodan Zhu$^{\S\,\ddagger}$, Zhenhua Ling$^\dagger$, Si Wei$^*$, Hui Jiang$^\ddagger$ \\ 
$^\dagger$University of Science and Technology of China, Hefei, China  \\
$^\S$National Research Council Canada, Ottawa, Canada \\
$^*$iFLYTEK Research, Hefei, China \\
$^\ddagger$York University, Toronto, Canada \\
\texttt{cq1231@mail.ustc.edu.cn, zhu2048@gmail.com, } \\
\texttt{zhling@ustc.edu.cn, siwei@iflytek.com, hj@cse.yorku.ca}}

\begin{document}

\maketitle

\begin{abstract}
Distributed representation learned with neural networks has recently shown to be effective in modeling natural languages at fine granularities such as words, phrases, and even sentences. Whether and how such an approach can be extended to help model larger spans of text, e.g., documents, is intriguing, and further investigation would still be desirable. This paper aims to enhance neural network models for such a purpose. A typical problem of document-level modeling is automatic summarization, which aims to model documents in order to generate summaries. In this paper, we propose neural models to train computers not just to pay attention to specific regions and content of input documents with attention models, but also distract them to traverse between different content of a document so as to better grasp the overall meaning for summarization. Without engineering any features, we train the models on two large datasets. The models achieve the state-of-the-art performance, and they significantly benefit from the distraction modeling, particularly when input documents are long.
\end{abstract}

\section{Introduction}

Modeling the meaning of text lies in center of natural language understanding. Distributed representation learned with neural networks has recently shown to be effective in modeling fine granularities of text, including words~\cite{collobert2011natural,mikolov2013distributed,Chen2015ACL}, phrases~\cite{yin-schutze:2014,Zhu:2014:ACL}, and arguably sentences~\cite{Socher2012,Irsoy2014,Kalchbrenner2014,Tai2015,Zhu2015,Zhu:2015:StarSem,Chen:2016:NLI,Zhu:2016:NAACL}. 

Whether and how such an approach can be extended to help model larger spans of text, e.g., documents, is intriguing, and further investigation would still be desirable, although there has been interesting research conducted recently along this line~\cite{Li2015AHN,Hu2015LCSTSAL,Wang2015LargerContextLM,hermann2015teaching}. A typical problem of document-level modeling is automatic summarization~\cite{Mani_Automatic01,Das07asurvey,Nenkova2013}, in which computers generate summaries for documents, based on their shallow or deep understanding of the documents. 

If one regards the process of representing input documents, generating summaries, and the interaction between them to be a (complicated) function, fitting such a function could expect to have a large-scale annotated dataset to estimate a large set of parameters, while on the other hand, hard-coding summarization knowledge (in different forms) or limiting the number of model parameters (e.g., as in many extractive summarization models) are often adopted when there are no enough training data. This work explores the former direction and utilizes relatively large datasets~\cite{Hu2015LCSTSAL,hermann2015teaching} to train neural summarization models. In general, neural networks, as universal approximators, can fit very complicated functions and have shown to be very effective on many problems recently. 

Understanding the input documents and generating summaries are both challenging. 
On the understanding side, much recent work seems to have suggested that distributed representations (often vectors) by themselves may not be adequate for representing sentences, let along with longer documents. Additional modeling such as soft or hard \textit{attention} has been applied to retrospect subsequences or even words in input text to remedy the limits, which has shown to improve performances of different tasks such as those discussed in~\cite{Bahdanau2014NeuralMT,Luong2015EffectiveAT,Rush2015ANA} among others. We regard this to be a mechanism that provides a connection between input document modeling (encoding) and summary generating (decoding), which could model a level of cognitive controls---human summarizers themselves often move between the input documents and summaries when they summarize a document. 

We consider this control layer to be important, and in this paper we focus on better designing this control layer for summarization. We propose neural models to train computers not just to pay attention to specific regions and content of input documents with attention models, but also distract them to traverse between different content of a document so as to better grasp the overall meaning for summarization. 

Without engineering any features, we train the models with two large datasets. The models achieve the state-of-the-art performance and  they significantly benefit from the distraction modeling, particularly when input documents are long. We also explore several technologies that have been applied to sentence-level tasks and extend them to document summarization, and we present in this paper the technologies that showed to help improve the summarization performance. Even when it is applied onto the models that have leveraged these technologies, the distraction models can further improve the performance significantly. In general, our models here aim to perform \textit{abstractive} summarization.

\section{Related work}
\label{sec:related}

\noindent \textbf{Distributed representation}
Distributed representation has shown to be effective in modeling fine granularities of text as 
discussed above. Much recent work has also attempted to model longer spans of text with neural networks~\cite{Li2015AHN,Hu2015LCSTSAL,lin2015hierarchical,Wang2015LargerContextLM,hermann2015teaching}. This includes research that incorporates document-level information for language modeling~\cite{Wang2015LargerContextLM,lin2015hierarchical} and that answers  questions~\cite{hermann2015teaching} by comprehending input documents with attention-based models. More relevant to ours, the work of ~\cite{Li2015AHN} learned distributed representation for short documents with the averaged length of about a hundred word tokens, although the objective is not  summarization. Summarization typically faces documents longer than those, and summarization may be  more necessary when documents are long. In this paper, we propose neural models for summarizing typical news articles with up to thousands of word tokens. We find it is necessary to enable computers not just to pay attention to specific content of input documents with attention models, but also distract them to traverse between different content so as to better grasp the overall meaning for summarization, particularly when documents are long. 

\noindent \textbf{Neural summarization models}
Automatic summarization has been intensively studied for both text ~\cite{Mani_Automatic01,Das07asurvey,Nenkova2013} and speech ~\cite{Zhu:2006:NAACL,Zhu:2009:SUMM}. Most state-of-the-art summarization models have focused on \textit{extractive} summarization, although some efforts have also been exerted on \textit{abstractive} summarization. Recent neural summarization models include the recent efforts of ~\cite{Rush2015ANA,Lopyrev2015GeneratingNH,Hu2015LCSTSAL}. The research performed in ~\cite{Rush2015ANA} focuses on neural models for sentence compression and rewriting, but not full document summarization. The work of ~\cite{Lopyrev2015GeneratingNH} leverages neural networks to generate news headline, where input documents are limited to 50 word tokens, and the work of~\cite{Hu2015LCSTSAL} also deals with short texts (up to dozens of word tokens), in which summarization problems such as content redundancy is less prominent and attention-based models seem to be sufficient. However, summarization typically faces documents longer than that and summarization is often  more needed when documents are long. In this work we attempt to explore neural summarization technologies for news articles with up to thousands of word tokens, in which we find distraction-based summarization models help improve performance. Note that our improvement is achieved over the model that has already outperformed the attention-based model reported in~\cite{Hu2015LCSTSAL} on short documents.

\section{Our approach}

\subsection{Overview}

We base our model on the general encoder-decoder framework ~\cite{ICML2011Sutskever_524,sutskever2014sequence,Cho2014LearningPR} that has shown to be effective recently on different tasks. This is a general sequence-to-sequence modeling framework in which the encoding part can be devoted to model the input documents and the decoder to generate output.

We believe the control layer that helps navigate the input documents to optimize the generation objectives would be of importance, and we will focus on the control layer in this paper and enrich its expressiveness. Specifically for summarization, unlike much recent work that focuses more on \textit{attention} in order to grasp local context or correspondence (e.g, in machine translation and sentence compression), we force our models to traverse between different content of a document to avoid focusing on a region or same content, to better grasp the overall meaning for the summarization objective. 

We also explore several popular technologies that have been applied to sentence-level tasks and extend them to document summarization, and we present those that help improve the summarization performance. 

\subsection{GRU-based encoding and decoding}

\noindent \textbf{Encoding} 
The general document modeling and summarizing framework takes in an input document $\vect{x}=x_1, \cdots,x_{\!\mathsmaller{T_x}}$ and write the summary of the document as the output $\vect{y}=y_1,\cdots,y_{\!\mathsmaller{T_y}}$. The summarization process is modeled as finding the output text $\vect{y}^*$ that maximizes the conditional probability  $\arg\max_{\vect{y}} p(\vect{y}|\vect{x})$, given gold summary sequences. As discussed above, such a model has been found to be very effective in modeling sentences or sub-sentential text spans. We will address the challenges faced at the document level.

On encoding we do not restrict the encoders' architectures as if it is a recurrent neural network (RNN). The recent literature shows long-short term memory (LSTM)~\cite{hochreiter1997long,sutskever2014sequence} and gated recurrent units (GRU)~\cite{Bahdanau2014NeuralMT} are both good architectures. In developing our systems we empirically found GRU achieved similar performance as LSTM but it is fast to train; we will therefore describe the GRU implement of our neural summarization models. 

In the simplest uni-directional setting, when reading input symbols from left to right, a GRU learns the hidden annotations $h_i$ at time $i$ with 
\begin{equation}
h_i=\textrm{GRU}(h_{i-1}, e(x_i))
\label{equ:GRU}
\end{equation}
\noindent where the $h_i \in \RR^n$ encodes all content seen so far at time $i$ which is computed from $h_{i-1}$ and $e(x_i)$, where $e(x_i) \in \RR^m$ is the $m$-dimensional embedding of the current word $x_i$. The forward propagation of GRU is computed as follows.
\begin{align}
\label{equ:inputH}
h_i &= (1 - u_i) \odot h_{i-1} + u_i \odot \tilde{h_i} \\[0ex]
\tilde{h_i} &= \tanh(W e(x_i) + U (r_i \odot h_{i-1}))\\[0ex]
r_i &= \textrm{sigmoid}(W_r e(x_i) + U_r h_{i-1})\\[0ex]
u_i &= \textrm{sigmoid}(W_u e(x_i) + U_u h_{i-1})
\end{align}
\noindent where $W_u$, $W_r$,  $W \in \RR^{n \times m}$ and $U_u$, $U_r$, $U \in \RR^{n \times n}$ are weight matrices, $n$ is the number of hidden units, and $\odot$ is element-wise multiplication. 

\begin{figure}[htp]
	\centering
	\includegraphics[width=\columnwidth]{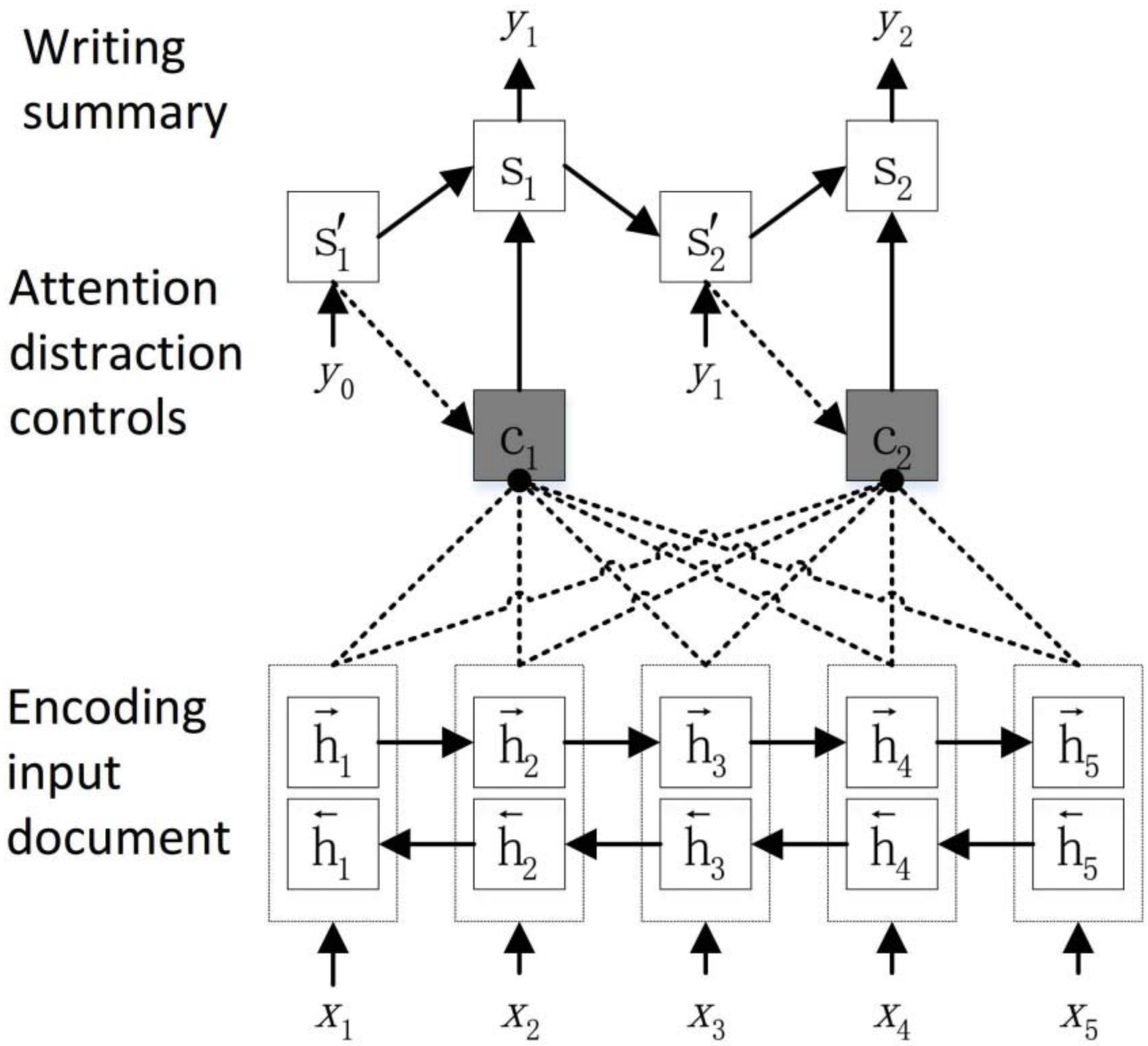}
	\caption{A high-level view of the summarization model. }
	\label{fig:model}
\end{figure}

In our work, we actually applied bi-directional GRUs (bi-GRUs), which we found achieving better results than single directional GRUs consistently. As its name suggests, in a bi-GRU unit, the annotation vector $h_t$ encodes the sequence from two directions, modeling both the left and right context. The bottom part of Figure~\ref{fig:model} shows the encoder intuitively, while for more details, readers can refer to ~\cite{Bahdanau2014NeuralMT} for further discussion. 

\noindent \textbf{Generation}
When generating summaries, the decoder predicts the next word $y_{t}$ given all annotations obtained in encoding $\vect{h}={h_1,\cdots,h_{\!\mathsmaller{T_x}}}$, as well as all previously predicted words ${y_1,\cdots,y_{t-1}}$. The objective is a probability over the summary $\vect{y}$ with decomposition into the ordered conditionals:
\begin{align}
\label{equ:obj}
\vect{y}^* &= \argmax_{\vect{y}} \prod_{t=1}^{\!\mathsmaller{T_y}} p(y_t|{y_1,\cdots,y_{t-1}},\vect{h})\\
& =\argmax_{\vect{y}} \prod_{t=1}^{\!\mathsmaller{T_y}} g(y_{t-1},s_t,c_t)|_{y_{t}} 
\label{equ:g}
\end{align}
\noindent where Equation~(\ref{equ:obj}) depicts a high level abstraction of generating a summary word $y_{t}$ over previous output as well as input annotations $\vect{h}=h_1,\cdots,h_{\!\mathsmaller{T_x}}$, and $y_t$ is a legal output word at time $t$, while $\vect{y}^*$ is the optimal summary found by the model.

The conditional probability is further rewritten as Equation~(\ref{equ:g}) to factorize the model according to the structures of neural networks. The function $g(y_{t-1},s_t,c_t)$ is a nonlinear function that computes the probability vector for all legal output words at output time $t$, and $g(y_{t-1},s_t,c_t)|_{y_t}$ takes the element of the resulting vector corresponding to word $y_t$, i.e., the predicated probability for word $y_t$. 

The vector $s_t$ and $c_t$ are the control layers that connect output $\vect{y}$ and input $\vect{h}$, which we will discuss in details in Section~\ref{subsec:cognitive}. For completeness, 
function $g(.)$ is computed with:
\begin{equation}
\Resize{7.6cm}{g(y_{t-1},s_t,c_t)=\sigma(W_o \tanh(V_o e(y_{t-1})+U_o{s}_t+C_o c_t))}
\label{equ:output}
\end{equation}
where $\sigma$ is a softmax function; $W_o \in \RR^{K \times n}$, $U_o \in \RR^{n \times n}$, $V_o \in \RR^{n \times m}$ and $C_o \in \RR^{n \times 2n}$ are weight matrices; $K$ is the vocabulary size; $e(y_{t-1}) \in \RR^m$ is the $m$-dimensional  embedding of the previously predicted word $y_{t-1}$.

\subsection{The control layers}
\label{subsec:cognitive}

The document modeling and summary generation are described above as two components: input document encoding and summary generation. A core problem is how these two components are associated. In sentence-level modeling such as machine translation and speech recognition, attention model is often applied to grasp local context and correspondence between input and output texts. For example, in translation attention is shown to be very useful for aligning the words of the target language (the language being translated to) to the corresponding source words and their context.

Attention can be regarded as a type of cognitive controls. In modeling documents, we take a general viewpoint on this control layer and propose distraction modeling to enable the model to traverse over different content of a long document, and we will show it improves the summarization performance significantly. 
In general, the control layer allows a complicated examination over the input. 
In this section we describe the controls that consider both attention and distraction to navigate input documents and to generate summaries.

\noindent \textbf {Two-layer hidden output}
We first extended the recent two-level hidden output model~\cite{Luong2015EffectiveAT} to our summarization models. As presented later in the experiments, the two-level hidden output model consistently improves the summarization performance on different datasets. More specifically, the updating of $s_t$ follows a two-layer GRU architecture shown in the top part of Figure~\ref{fig:model}. 
\begin{align}
s_t&=\textrm{GRU}_1(s_t', c_t)\\
s_t'&=\textrm{GRU}_2(s_{t-1}, e(y_{t-1}))
\end{align}
The forward propagation of $\textrm{GRU}_1$ and $\textrm{GRU}_2$ are computed similar to Equation~(\ref{equ:GRU}) above.  $\textrm{GRU}_1$ and $\textrm{GRU}_2$ use untied parameter matrices. The two-layer model allows for capturing a direct interaction between $s_t'$ and $c_t$, with the former encoding the current and previous output information and the latter encoding the current input content that is primed with distraction and attention. We will discuss how these vectors are computed below. 

\noindent \textbf{Distraction in training} 
We propose to enforce distraction from two perspectives: adding the distraction constraints in training as well as in decoding. We first discuss the distraction in training.

\noindent \textit{\underline{\smash{Distraction over input content vectors}}}
In training we force the model not to pay attention to the same content or same part of the input documents too much. We accumulate the previously viewed content vector as a \textit{history} content vector $\sum_{j=1}^{t-1} c_{j}$ and incorporate it into the currently computed $c_t'$. We refer to this model as \textbf{M1}. 
\begin{equation}
c_t=\tanh (W_c {c}_t' - U_c \sum_{j=1}^{t-1} c_{j})
\label{equ:c}
\end{equation}
\noindent where $W_c \in \RR^{2n \times 2n}$ and $U_a \in \RR^{2n \times 2n}$ are diagonal matrices. And $c_t'$ is  input content vectors that have not been directly penalized with history yet; $c_t'$ is directly computed with conventional equation as follows:
\begin{equation}
c_t'=\sum_{i=1}^{\!\mathsmaller{T_x}}\alpha_{t,i}h_i
\label{equ:c1}
\end{equation}
\noindent where $h_i$ are annotation vectors that encode the current input word and its context with the input GRU described above in Equation~(\ref{equ:GRU}). And $\alpha_{t,i}$ is the attention weight put on $h_i$ at the current output time $t$.
The distraction-based $c_t$ computed in Equation~(\ref{equ:c}) can then be incorporated in Equation~(\ref{equ:output}). 

\noindent \textit{\underline{\smash{Distraction over attention weight vectors}}}
We also propose to add distraction directly on the attention weight vectors. Similarly as above, we accumulate the past attention weights as a history attention weight $\sum_{j=1}^{t-1}\alpha_{j,i}$ and use it to prime the current attention weights. The model in ~\cite{Tu2016CoveragebasedNM} also uses history attention weights, but we use history here to force distraction in order to avoid redundancy, which is not a concern in the machine translation task. We refer to the model as~\textbf{M2}. 
\begin{equation}
\label{equ:align}
\alpha_{t,i}'=v_a^T\tanh(W_a s_t'+U_a h_i - b_a \sum_{j=1}^{t-1}\alpha_{j,i})
\end{equation}
\noindent where $W_a \in \RR^{l \times n}$, $U_a \in \RR^{l \times 2n}$, $b_a \in \RR^{l}$, and $v_a \in \RR^{l}$ are the weight matrices, and $l$ is the number of hidden units. Note that $W_a s_t'+U_a h_i$ in the equation computes the conventional attention without penalizing attention history. $\alpha'$ is often normalized with a softmax to generate attention weights $\alpha_{t,i}$ below, which is in turn used in Equation~(\ref{equ:c1}).  
\begin{equation}
\label{equ:weight}
\alpha_{t,i}=\frac{\exp(\alpha_{t,i}')}{\sum_{j=1}^{T_x}\exp(\alpha_{t,j}')}
\end{equation}

\noindent \textbf{Distraction in decoding}
In the decoding process, we also enforced different types of distraction, one by computing the difference between the distribution of the current attention weight $\alpha_t$ and that of all previous attention weights $\alpha_1,\cdots,\alpha_{t-1}$. Since $\alpha$ can be seen as a proper probabilistic distribution, normalized in Equation~(\ref{equ:weight}), we used Kullback-Leibler (KL) divergence to measure their difference with Equation~(\ref{equ:alphaDiff}), which was found to be consistently better than several other distance metrics we tried on the held-out data. 

We also enforced distraction in a similar way on the attention-primed input content vector $c_t$, as well as on the hidden output vector $s_t$. Both $c_t$ and $s_t$ are not normalized but are regular content vectors, where the cosine similarity was found achieving a better performance than several alternatives (e.g., $l_1$ and $l_2$ distances) on the held-out data. 
\begin{align}
\label{equ:alphaDiff}
d_{\alpha,t} &= \min_{i \in \{1, \cdots t-1\}} \textrm{KL}(\alpha_t, \alpha_i)\\
d_{c,t} &= \max_{i \in \{1, \cdots t-1\}} \textrm{cosine}(c_t, c_i)\\
d_{s,t} &= \max_{i \in \{1, \cdots t-1\}} \textrm{cosine}(s_t, s_i)
\end{align}
\noindent The distraction score $d_{*,t}$ was then added into the output probability and the beam search in order to encourage the model to avoid redundant content. 
\begin{equation}
\textrm{score}_t=\sum_{t=1}^{\!\mathsmaller{T_y}}\left\{\log(p_{y_t}) + \lambda_1 d_{\alpha,t} + \lambda_2 d_{c,t} + \lambda_3 d_{s,t}\right\}
\end{equation}
\noindent where $score_t$ was used as follows in the \textit{beam search with distraction} algorithm, and parameter $\lambda_1$, $\lambda_2$, and $\lambda_3$ were determined on the development set. We refer to this model as~\textbf{M3}.

\begin{algorithm}[ht]
	\begin{algorithmic}
		\Require Vocabulary size $K$, beam size $B$, max output length $N$
		\LineComment{Computed probabilities of all the words in vocabulary}
		\LineComment{Choose the $B$ most likely words and initialize the $B$ hypotheses}
		\For{$i=1$ : $N$}
		\LineComment{For each hypothesis, compute the next conditional probabilities, then have $B \times K$ candidates with the corresponding probabilities}
		\LineComment{Use the distraction-primed value \textit{score} to choose $B$ most likely candidates}
		\EndFor
	\end{algorithmic}
	\caption{Beam search with distraction}
	\label{alg:beam}
\end{algorithm}

\noindent \textbf{Unknown word replacement for summarization}
We borrowed the unknown word replacement~\cite{Jean2015OnUV} from machine translation to our summarization models and found it improved the performance when summarizing long documents. Specifically, due to the time complexity in handling a larger vocabulary in the softmax layer in summary generation, infrequent words were removed from the vocabulary and were replaced with the symbol $\langle {U\!N\!K} \rangle$. The threshold of vocabulary size is data-dependent and will be detailed later in the experiment set-up section.

After the first-round summary generated for a document, a  token labeled as $\langle {U\!N\!K} \rangle$ will be replaced with a word in the input documents. More specifically, we obtained the replacement using Equation~(\ref{equ:weight}); i.e., we used the largest element in $\alpha_t$ to find the source location for the current $\langle {U\!N\!K} \rangle$. 

\section{Experiment set-up}

\subsection{Data}
We experiment with our summarization models on two publicly available corpora with different document lengths and in different languages: a CNN news collection ~\cite{hermann2015teaching} and a Chinese corpus made available more recently in~\cite{Hu2015LCSTSAL}. Both are large datasets appropriate for training neural models, which, as discussed above, employ a large number of parameters to fit the potentially complicated summarization process  involving representing input documents,  generating summaries, and interacting between them. 

\noindent \textbf{CNN data}
The CNN data ~\cite{hermann2015teaching} have a human-generated real-life summary for each news article. The dataset collected in was made available at GitHub.\footnote{{https://github.com/deepmind/rc-data}} The data was preprocessed with the Stanford CoreNLP tools~\cite{manning2014stanford} for tokenization and sentence-boundary detection; all capital information is kept. To speed up training, we removed the documents that are too long (over 2,500 word tokens) from the training and validation set, but kept all documents in the test set, which does not change the difficulty of the task. 

\noindent \textbf{LCSTS data} The second corpus is LCSTS, which is a Chinese corpus made available more recently in~\cite{Hu2015LCSTSAL}. The data is constructed from the Chinese microblogging website, Sina Weibo. We used the original training/testing split mentioned in ~\cite{Hu2015LCSTSAL}, but additionally randomly sampled a small part of the training data as our validation set. 

Table~\ref{statistics} gives more details about the two datasets. We can see from the table that averaged document length of the CNN corpus is about seven time as long as the LCSTS corpus, and the summary is about 2-3 times longer.

\begin{table}[th!]
	\begin{center}
		\renewcommand{\arraystretch}{1}
		\setlength\tabcolsep{1.5pt}
		\begin{tabular}{|l|rrr|rrr|}
			\hline 
			&\multicolumn{3}{c|}{\bf CNN}&\multicolumn{3}{c|}{\bf LCSTS}\\
			&Train&Valid&Test&Train&Valid&Test\\
			\hline
			\hline
			Doc.L. & 775.2 & 761.3 & 765.4 & 103.7 & 100.3 & 108.1 \\
			Sum.L. & 48.4 & 36.5 & 36.6 & 17.9 & 18.2 & 18.7\\
			\hline 
			\hline 
			\# Doc. &81,824&1,184&1,093&2,400,000&591&725\\ 
			\hline
		\end{tabular}
	\end{center}
	\caption{\label{statistics} The CNN and LCSTS dataset. The first two rows of the table are the averaged document length (Doc.L.) and summary length (Sum.L.) in terms of numbers of word tokens. The bottom row lists the number of documents in the datasets.}
\end{table}

\subsection{Training details}

We used mini-batch stochastic gradient descent (SGD) to optimize log-likelihood, and Adadelta~\cite{Zeiler2012ADADELTAAA} to automatically adapt the learning rate of parameters ($\epsilon=10^{-6}$ and $\rho=0.95$). 

For the CNN dataset, training was performed with shuffled mini-batches of size 64 after sorting by length. We limit our vocabulary to include the top 25,000 most frequent words. Other words were replaced with the token $\langle {U\!N\!K} \rangle$, as discussed earlier in the paper. Based on the validation data, we set embedding dimension to be 120, the vector length in hidden layers to be 500 for uni-GRU and 600 for bi-GRU. An end-of-sentence token was inserted between every sentence, and an end-of-document token was added at the end. The beam size of decoder was set to be 5.

For the LCSTS data, a larger mini-batch size 256 was found to be better based the observation on the validation set. Same as in ~\cite{Hu2015LCSTSAL}, we used characters rather than words as our tokens. The vocabulary size is 4000, embedding dimension is 500, and the vector size of the hidden-layer nodes is 500. Beam search size is 5, same as in the CNN dataset. 

We make our code publicly available\footnote{Our code is available at https://github.com/lukecq1231/nats}. Our implementation uses python and is based on the Theano library~\cite{bergstra2010theano}.

\section{Experimental results}

\subsection{Results on the CNN dataset}

\noindent \textbf{Overall performance}
Our results on the CNN dataset are presented in Table~\ref{CNN_result}. We used Rouge scores~\cite{Lin2004ROUGEAP} to measure performance. Since the summary lengths are not preset to be the same, we report $F_1$ Rouge. The upper part of the table includes the baseline results of a number of typical summarization algorithms, which we listed in the table as Luhn~\cite{luhn1958automatic}, Edmundson~\cite{edmundson1969new},
LSA~\cite{steinberger2004using}, Lex-rank~\cite{erkan2004lexrank},
Text-rank~\cite{Mihalcea2004TextRankBO},
Sum-basic~\cite{vanderwende2007beyond}, and
KL-sum~\cite{haghighi2009exploring}. These baseline results are implemented in the open-source tool SUMY\footnote{ https://pypi.python.org/pypi/sumy}.  

The results at the lower half of the table show that the bi-GRU encoder achieves a better performance than the uni-GRU encoder. This is consistent with the results on the LCSTS dataset reported later in Table~\ref{tab:result_LCSTS}. We show that two-level output model we discussed in the method section is beneficial, which is also consistent with the results on the LCSTS dataset. In addition, the unknown replacement technique yields an additional improvement. 

Over the strong model that has used these technologies (the row marked as "+UNK replace"), the model in the last row that incorporates all distraction modeling (M1, M2 and M3) finally achieves a Rouge-1 score of 27.1, a Rouge-2 score of 8.2, and a Rouge-L score of 18.7, significantly improving the three Rouge scores by 5.8, 1.9, and 2.3, respectively. These are also the largest improvement presented in the table, compared with the other techniques listed. The table also lists the details of how the model M1, M2, and M3 improve the performance additively. Again, the neural models do not engineer any features and use only content but not any additional formality features such as locations of input sentences, which may bring additional improvement.

\begin{table}[h!]
	\begin{center}
		\renewcommand{\arraystretch}{1}
		\setlength\tabcolsep{2.8pt}
		\begin{tabular}{|l|ccc|}
			\hline 
			System&Rouge-1&Rouge-2&Rouge-L\\
			\hline
			\hline
			Luhn&23.2&7.2&15.5\\
			Edmundson&24.5&8.2&16.7\\
			LSA&21.2&6.2&14.0\\
			Lex-rank&26.1&\textbf{9.6}&17.7\\
			Text-rank&23.3&7.7&15.8\\
			Sum-basic&22.9&5.5&14.8\\
			KL-sum&20.7&5.9&13.7\\
			\hline
			\hline
			Uni-GRU&18.4 & 4.8 & 14.3 \\
			Bi-GRU&19.5 & 5.2 & 15.0 \\
			+Two-level out &20.2&5.9&15.7 \\
			+UNK replace&21.3 & 6.3& 16.4\\
			+Distraction M1& 22.2 & 6.5 & 16.7\\
			+Distraction M2& 24.4 & 7.7 & 17.8\\
			+Distraction M3&\textbf{27.1}&8.2&\textbf{18.7}\\
			\hline
		\end{tabular}
	\end{center}
	\caption{\label{CNN_result} Results on the CNN dataset.}
\end{table}

\noindent \textbf{Performance on different lengths of documents} To observe the effectiveness of the distraction model over different document lengths, we further selected all short documents from the CNN training dataset into a subset (subset-1) with average length at 335 word tokens, and a subset of data (subset-2) that have the same number of documents as the subset-1, with averaged document length at 680 word tokens.  As shown in Table~\ref{tab:diffSize}, on the data subset-2, the distraction model improves the results more significantly. The relative improvement is 29.0\%, 25.6\%, and 10.8\%, compared with 25.9\%, 20.5\%, and 8.1\% on subset-1, respectively. 
In general the best performance on both dataset is lower than that using all training data, suggesting using more training data can improve summarization performance. 

\begin{table}[h]
	\begin{center}
		\renewcommand{\arraystretch}{1}
		\setlength\tabcolsep{3.8pt}
		\begin{tabular}{|l|ccc|}
			\hline 
			&Rouge-1&Rouge-2&Rouge-L\\
			\hline
			\hline
			&\multicolumn{3}{c|}{Subset-1}\\
			\hline
			w.o. distraction&{18.1}&{3.9}&{13.6} \\
			+Distraction&{22.8}&{4.7}&{14.7} \\
			Relative Impr.&{25.9\%}&{20.5\%}&{8.1\%} \\
			\hline
			\hline
			&\multicolumn{3}{c|}{Subset-2}\\
			\hline
			w.o. distraction&{18.6}&{3.9}&{13.8} \\
			+Distraction&{24.0}&{4.9}&{15.3} \\
			Relative Impr.&{29.0\%}&{25.6\%}&{10.8\%} \\
			\hline
		\end{tabular}
	\end{center}
	\caption{\label{tab:diffSize} Results on two subsets of the CNN datasets with different document lengths. }
\end{table}

\subsection{Results on the LCSTS dataset}

We experiment with the proposed model on public LCSTS corpus. The baseline is the best result reported in~\cite{Hu2015LCSTSAL}\footnote{We thank the authors of ~\cite{Hu2015LCSTSAL} for generously sharing us the latest output of their models, which achieves a better performance than the results reported in ~\cite{Hu2015LCSTSAL}. We reported here the updated scores higher performance as our baseline.}. Our modified uni-GRU achieves a slight improvement over the reported results. The Bi-GRU attention-based model achieves a better performance, confirming the usefulness of bi-directional models for summarization as well as that our implementation is the state-of-the-art and serves as a very strong baseline in the CNN dataset discussed above. Note that since the input text length of LCSTS is far shorter than the CNN documents, each containing about 100 words and roughly 6-8 sentences, we show that distraction does not improve the performance, but in contrast, when documents are longer, its benefits are significant, achieving the biggest improvement as discussed earlier. This suggests the effectiveness of  distraction modeling in helping summarize the more challenging longer documents, where summarization is often more necessary than for short texts. 

\begin{table}[h]
	\begin{center}
		\renewcommand{\arraystretch}{1}
		\setlength\tabcolsep{3.8pt}
		\begin{tabular}{|l|ccc|}
			\hline 
			System&Rouge-1&Rouge-2&Rouge-L\\
			\hline
			\hline
			\cite{Hu2015LCSTSAL}&29.9&17.4&27.2\\
			Uni-GRU &32.1&19.9&29.4 \\
			Bi-GRU &33.2&20.8&30.5\\
			+Two-level Att.&35.2&22.6&{32.5} \\
			+UNK replace&{35.2}&{22.6}&{32.5} \\
			+Distraction&{35.2}&{22.6}&{32.5} \\
			\hline
		\end{tabular}
	\end{center}
	\caption{\label{tab:result_LCSTS} Results on the LCSTS dataset. }
\end{table}

We also compare our models with the simple baseline that selects the first $N$ numbers of word tokens from the input documents, which reaches its maximal Rouge scores when the first 30 tokens were taken, and achieves Rouge-1, Rouge-2, and Rouge-L at 25.5, 14.1 and 21.4. And our models are significantly better than that. For the CNN data set, choosing the first three sentences achieves the best results, which reach Rouge-1, Rouge-2, and Rouge-L at 26.1, 9.6 and 17.8, respectively. Since the CNN data is news data, the baseline of selecting first several sentences has known to be a very strong baseline. Again, the models we explore here are towards performing  abstractive summarization.

\section{Conclusions and future work}

We propose to train neural document summarization models not just to pay attention to specific regions of input documents with attention models, but also distract the models to different content in order to better grasp the overall meaning of input documents. Without engineering any features, we train the models on two large datasets. The models achieve the state-of-the-art performance and  they significantly benefit from the distraction modeling, particularly when the input documents are long. We also explore several recent technologies for summarization and show that they help improve summarization performance as well. Even if applied onto the models that have already leveraged these technologies, the distraction models can further improve the performance significantly.

From a more general viewpoint, enriching the expressiveness of the control layers that link the input encoding layer and the output decoding layer could be of importance to remedy the shortcomings of the current models. We plan to perform more work along this direction. 

\section*{Acknowledgments}

\noindent The first and the third author of this paper were supported in part by the Science and Technology Development of Anhui Province, China (Grants No. 2014z02006), the Fundamental Research Funds for the Central Universities (Grant
No. WK2350000001) and the Strategic Priority Research Program of the Chinese Academy of Sciences (Grant No. XDB02070006).

\bibliographystyle{named}
\small\bibliography{ijcai16}

\end{document}